\pdfoutput=1

\documentclass[11pt]{article}

\usepackage{ACL2023}

\usepackage{times}
\usepackage{latexsym}

\usepackage[T1]{fontenc}

\usepackage[utf8]{inputenc}

\usepackage{microtype}

\usepackage{inconsolata}

\usepackage{amssymb}
\usepackage{breakurl}
\usepackage{booktabs}
\usepackage{multirow}
\usepackage{comment}
\usepackage{pifont}
\usepackage[textsize=scriptsize]{todonotes}

\usepackage[final]{draftwatermark}
\SetWatermarkText{THIS IS A DRAFT}
\SetWatermarkScale{2}
\SetWatermarkColor[gray]{0.9}
\SetWatermarkAngle{30}
\usepackage[whole]{bxcjkjatype}
\usepackage{subcaption}
\usepackage{ascmac}
\usepackage{fancybox}
\usepackage{gb4e}
\usepackage{cgloss4e}
\noautomath

\newcommand{\threetimes}{\textbf{3\_times}}
\newcommand{\seventimes}{\textbf{7\_times}}
\newcommand{\zeroshot}{\textbf{zero-shot}}
\newcommand{\simple}{\textsc{simple}}
\newcommand{\complex}{\textsc{complex}}
\newcommand{\simpletrain}{\textsc{simple\_train}}
\newcommand{\simpletest}{\textsc{simple\_test}}
\newcommand{\complextest}{\textsc{complex\_test}}
\newcommand{\nominative}{\textsc{nom}}
\newcommand{\dative}{\textsc{dat}}
\newcommand{\accusative}{\textsc{acc}}
\newcommand{\citative}{\textsc{cite}}
\newcommand{\citeau}[1]{\citeauthor{#1},~\citeyear{#1}}
\newcommand{\wraptext}[1]{\begin{tabular}{@{}l}#1\end{tabular}}

\title{Analyzing Syntactic Generalization Capacity of Pre-trained Language Models on Japanese Honorific Conversion}

\author{Ryo Sekizawa \and Hitomi Yanaka \\
    The University of Tokyo \\
    \texttt{\{ryosekizawa,hyanaka\}@is.s.u-tokyo.ac.jp}}

\begin{document}

\maketitle

\begin{abstract}
Using Japanese honorifics is challenging because it requires not only knowledge of the grammatical rules but also contextual information, such as social relationships.
It remains unclear whether pre-trained large language models (LLMs) can flexibly handle Japanese honorifics like humans. 
To analyze this, we introduce an honorific conversion task that considers social relationships among people mentioned in a conversation. 
We construct a Japanese honorifics dataset from problem templates of various sentence structures to investigate the syntactic generalization capacity of GPT-3, one of the leading LLMs, on this task under two settings: fine-tuning and prompt learning.
Our results showed that the fine-tuned GPT-3 performed better in a context-aware honorific conversion task than the prompt-based one.
The fine-tuned model demonstrated overall syntactic generalizability towards compound honorific sentences, except when tested with the data involving direct speech. 
\end{abstract}
\begin{table*}[t]
    \small\centering
    \begin{tabular*}{\textwidth}{lll} 
    \toprule
    Type & Target of respect and deference & Example \\    
    \midrule\midrule
    Subject honorifics (SH) & Agent & \underline{Sensei}-ga Hanako-o \textbf{homete-irasshatta}.\\   
    && Teacher-\textsc{nom} Hanako-\textsc{acc} praised-\textsc{sh}\\
    Object honorifics (OH) & Patient & (Watashi-ga) \underline{sensei}-no-tokoro-ni \textbf{ukagau}.\\
    && (I-\textsc{nom}) teacher-\textsc{poss}-place-\textsc{loc} visit-\textsc{oh}\\
    \bottomrule
    \end{tabular*}
    \caption{Types of Japanese honorifics with conjugation rules. The underlined part is a person to whom the speaker should show respect or deference. The bolded parts are conjugated verbs. A verb \textit{hometa} (praised) conjugates to its subject honorific form \textit{homete-irasshatta} by attaching a suffix \textit{irasshatta} and \textit{tazuneru} (visit) conjugates to its object honorific form \textit{ukagau}.}
    \label{table:cr_honorific_types}
\end{table*}
\begin{figure*}[t]
    \centering
    \includegraphics[width=\textwidth]{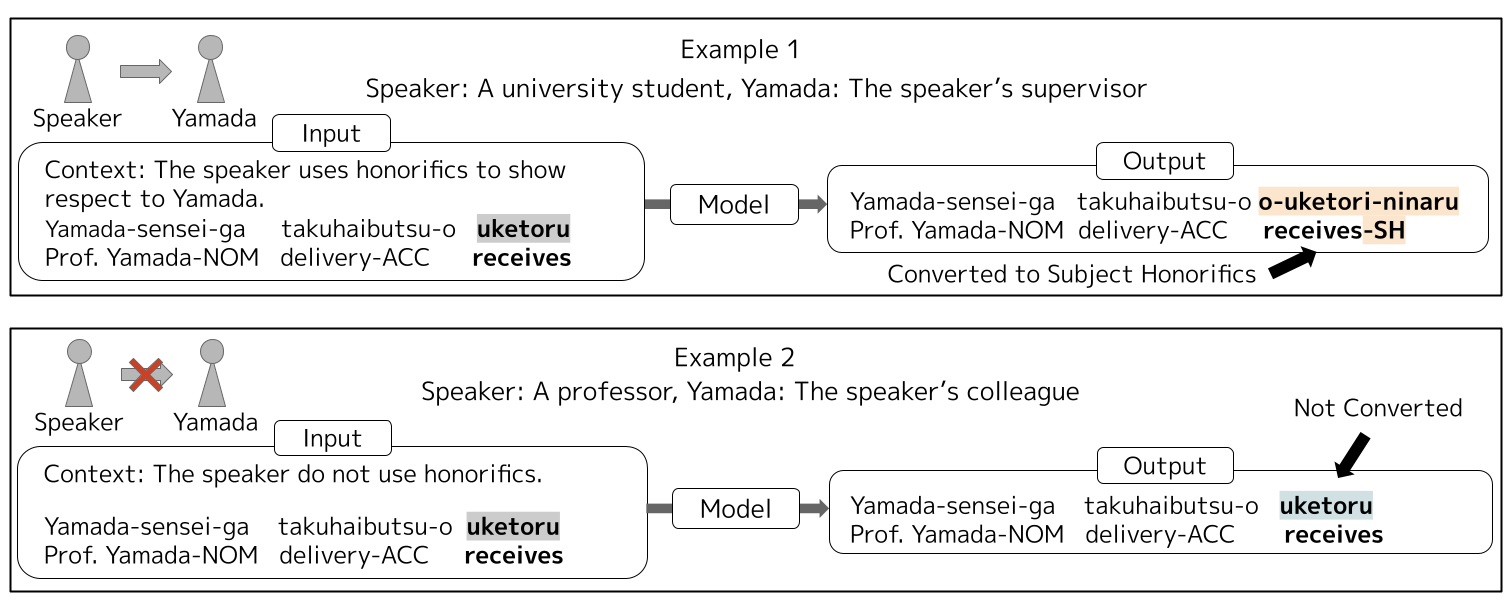}
    \caption{Examples of an honorific conversion task that considers contextual information. The bolded verbs conjugate to their honorific form if needed, considering the context.}
    \label{figure:cr_honorific_conversion}
\end{figure*}
\begin{figure}[t]
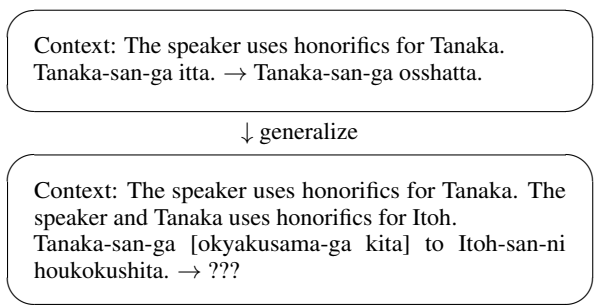

    \footnotesize\centering
    \begin{screen}
    Context: The speaker uses honorifics for Tanaka.\\
    Tanaka-san-ga itta. $\rightarrow$ Tanaka-san-ga osshatta.
    \end{screen}
    \centering$\downarrow$ generalize
    \begin{screen}
    Context: The speaker uses honorifics for Tanaka. The speaker and Tanaka uses honorifics for Itoh.\\
    Tanaka-san-ga [okyakusama-ga kita] to Itoh-san-ni houkokushita. $\rightarrow$ ???
    \end{screen}
    \caption{An example of the syntactic generalization of honorifics. The lower problem is made of a more complex sentence structure (center embedding and indirect speech) than the upper problem.}
    \label{figure:cr_syntactic_generalization}
    \vspace{-0.6cm}
\end{figure}
\section{Introduction}
\label{sec:cr_introduction}
The correct use of Japanese honorifics is difficult because it requires both the knowledge of grammatical rules (i.e., verb conjugation) and contextual information (i.e., social relationships among the speaker, the hearer, and the people mentioned in a conversation) ~\cite{harada_honorifics_1976}.
We expect this syntactic and pragmatic ability for pre-trained large language models (LLMs), as they have shown high performance on natural language tasks (\citeau{brown-gpt3-2020}, ~\citeau{ouyang_instructgpt_2022}, \textit{inter alia}).
However, it remains unclear whether LLMs can handle Japanese honorifics in a similar manner to humans, based on sentence structures and social context. 

Several studies proposed datasets of Japanese honorifics for classification~\cite{liu_construction_2022,someya_jcola_2022} and generation~\cite{matsumoto_honorifics-style-transfer_2022}.
~\citet{liu_construction_2022} introduced a task in which a model takes an honorific sentence as input and classifies its honorific level or the types of honorifics used in the sentence.
~\citet{someya_jcola_2022} provided a Japanese acceptability classification dataset called JCoLA.
In JCoLA, subject honorifics are categorized as sub-categories of subject-verb agreement tasks.
However, these datasets aim to evaluate the syntactic performance of language models, and they do not analyze their pragmatic ability to understand honorifics by considering social relationships behind sentences.
~\citet{matsumoto_honorifics-style-transfer_2022} introduced an evaluation dataset for an honorific conversion task in which the input was a non-honorific sentence, and the output was an honorific sentence.
~\citet{matsumoto_honorifics-style-transfer_2022} mentioned the necessity of considering the information on social relationships among people in honorific conversion but did not clarify how such information should be processed in the task.
In summary, the existing benchmark datasets of Japanese honorifics focus on either syntactic or pragmatic knowledge required for honorific understanding, not both (Appendix~\ref{sec:cr_appendix_existing_datasets}).
Additionally, none of these existing studies discusses the generalization capacity toward various syntactic structures of honorific sentences.

In this research, we introduce a new honorific conversion task that uses information on person's social relationships as additional input.
In~\citet{matsumoto_honorifics-style-transfer_2022}'s proposed honorific conversion, the input was only a non-honorific sentence.
In our task, social relationships are expressed as a sentence and  concatenated into an input sentence (Section~\ref{sec:cr_task_overview}).
This enables us to analyze whether LLMs could consider information on social relationships when executing honorific conversion.
We also construct a dataset to investigate the syntactic generalizability of LLMs for this honorific task. 
We create hand-crafted templates and generate problems for the task by filling in the placeholders (Section~\ref{sec:cr_dataset_construction}). 
By focusing on the syntactic generalization capacity, we analyze how flexibly LLMs can apply the grammatical rules of honorifics.
Using our dataset, we then fine-tune and evaluate the performance of GPT-3 on the task (Section~\ref{sec:cr_experiments}).
Additionally, we evaluate the models using zero-shot learning to determine how well these models perform for honorific conversion using the prompt-based method.
Our experiments indicate that the fine-tuned GPT-3 successfully generalizes to sentences with more complex structures, such as scrambling, but not to those involving direct speech.
We also show that the model with prompt learning demonstrates much lower performance than that with fine-tuning.

Our dataset will be publicly available at \texttt{\url{https://github.com/ynklab/japanese_honorifics}}.
\begin{figure*}[t]
    \centering
    \includegraphics[width=\textwidth]{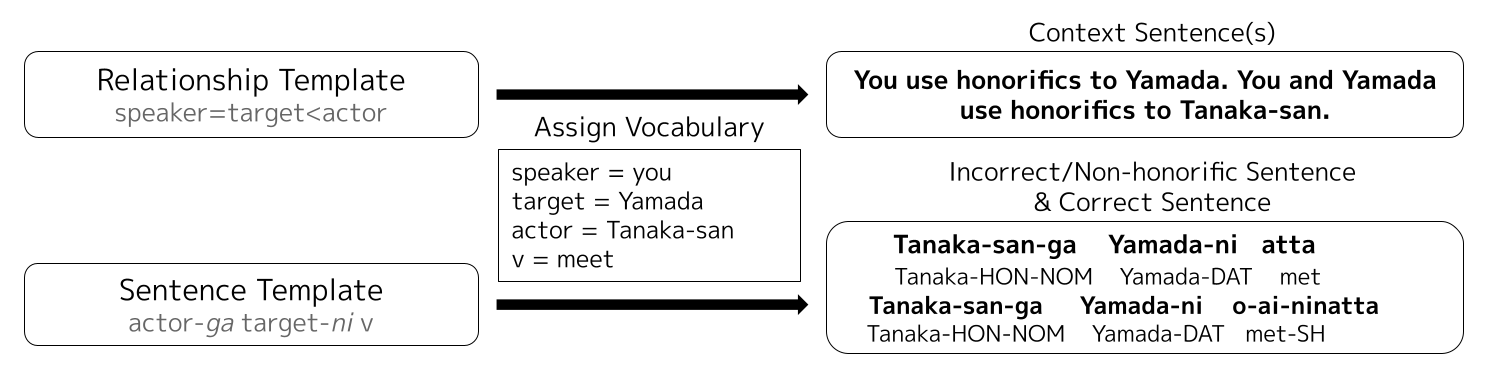}
    \caption{Overview of the process of dataset construction. The bolded sentences are used in the task. The verb \textit{met} has to conjugate to its subject honorific form \textit{met}-\textsc{sh} (\textit{o-ai-ni-natta}) since the speaker is supposed to use honorifics for \textit{Tanaka-san}.}
    \label{figure:cr_dataset_construction}
    \vspace{-0.5cm}
\end{figure*}

\section{Task Overview}
\label{sec:cr_task_overview}
\paragraph{Japanese Honorifics}
\label{par:japanese_honorifics}
Japanese honorifics are based on various linguistic phenomena~\cite{bunkacho_shishin_2007, kijutsu_japanese_2009}; some have grammatical rules of conjugation.
We target Subject Honorifics (SH) and Object Honorifics (OH).
As shown in Table~\ref{table:cr_honorific_types}, these honorifics are applied depending on the grammatical position of \textit{sensei} (a teacher) so that the speaker can express respect or deference towards the teacher.
SH is applied to the predicate when \textit{agent} has a higher social status than the speaker, and OH is applied when the \textit{patient} has a higher social status than the speaker.
\paragraph{Honorific Conversion}
\label{par:honorific_conversion}
The existing research proposed honorific conversion~\cite{matsumoto_honorifics-style-transfer_2022}.
We extend this task to include sentences explaining social relationships as input.
In the upper example of Figure~\ref{figure:cr_honorific_conversion}, the speaker is talking about supervisor Yamada's actions, so the verb \textit{uketoru} (receive) should be converted into the subject honorific form.
In the lower example, the speaker and Yamada are in a casual relationship because they are colleagues; therefore, the model should output the same sentence as the input without honorific conjugation.
\paragraph{Syntactic Generalization}
\label{par:syntactic_generalization}
We focus on the models' syntactic generalization ability to capture whether models can flexibly use honorific rules.
In this paper, syntactic generalization refers to a model's ability to use honorific rules for not only simple syntactic structures but also complex syntactic structures (see Figure~\ref{figure:cr_syntactic_generalization}).
\section{Dataset Construction}
\label{sec:cr_dataset_construction}
We construct a Japanese honorific dataset by manually creating problem templates and filling their placeholders with vocabulary using dictionaries to evaluate LLMs' performance on the honorific task.
Our dataset construction method is shown in Figure~\ref{figure:cr_dataset_construction}.
We take this approach instead of automatically collecting data from corpora for two reasons.
First, it is difficult to create sentence data with complex structures, such as scrambling, from corpora in a controlled manner.
This possibly makes it easier for a model to do honorific conversion than when the information is implicit.
Second, we need to prepare controlled settings for social relationship information to evaluate whether LLMs utilize it in honorific conversion; however, such information does not appear explicitly in the corpora.

The second problem is related to the fact that words in argument positions are often dropped in Japanese, especially in dialogue sentences.
(1) and (2) display a conversation between a junior worker and their boss.
\begin{exe}
    \ex \textit{A boss asks a question to a junior}
    \vspace{-0.2cm}
    \gll Okashi-tte mada nokotteiru ? \\
    snack-\textsc{top}, still remain ? \\
    \vspace{-0.2cm}
    \glt `Are there any snacks left?' \\
    \vspace{-0.3cm}
    \ex \textit{The junior answers}
    \vspace{-0.2cm}
    \gll $\phi_i$ nokori wa itadakimashita$_i$\\
    $\phi$ remained \textsc{top} had-\textsc{oh}\\
    \vspace{-0.2cm}
    \glt `I/We had them all.' \\
    \vspace{-0.5cm}
\end{exe}
\noindent In this conversation, two ambiguous points must be clarified to determine the honorific relationship behind the conversation.
The first point is that the junior answers with object honorifics to show deference in (2), but the target of deference is ambiguous without additional context.
If snacks are something the boss originally brought to their office, the boss is the target of the junior's deference.
However, if the snacks are prepared by some third person with a higher rank of job position than the junior, the deference must be towards them instead of the boss.
The second point is that the subject is dropped in (2) (pro-drop), but we cannot determine whether $\phi$ refers to the junior or to a group of workers containing the junior.
Considering such language-specific phenomena, we take a template-based approach instead of a corpus-based approach for dataset construction.
\subsection{Templates}
\label{sec:cr_creating_templates}
We create 39 problem templates based on the literature on Japanese linguistics~\cite{bunkacho_shishin_2007, kijutsu_japanese_2009}.
A graduate student with a linguistic background created the templates by consulting a linguistics researcher.
Each problem template has three elements for generating input and output data for honorific conversion: the relationship template, sentence template, and honorific type.
\paragraph{Relationship Templates}
\label{para:cr_relationship_templates}
Relationship templates represent social relationships among a speaker, a person who makes an action (agent), and one who is the target of the action (patient) in an equation-like format.
For example, \textit{speaker=actor<target} means that the speaker and actor do not use honorifics for each other and should use honorifics for the target.
\paragraph{Sentence Templates}
\label{para:cr_sentence_templates}
Sentence templates have placeholders for person's names and verbs.
Based on their structural complexities, we prepare two types of sentence templates: \simple{} and \complex{}.
\simple{} is a template that has one clause and S(O)V structure, and \complex{} is a template that has more complex syntactic structures: scrambling (SC), center embedding (CE), direct speech (DS), and indirect speech (IS). 
The first two structures change the argument positions within a sentence, potentially posing challenges for the model in capturing subject-verb agreement.
The last two are related to honorific application, depending on whether the sentence has quotation marks (brackets) or not (see Appendix~\ref{sec:cr_appendix_templates}).
A \complex{} template may contain multiple structures (e.g., IS \& CE).
See Appendix~\ref{sec:cr_appendix_templates} for further details.
\subsection{Problem Generation}
\label{sec:cr_problem_generation}
We create problem data for training and evaluating models by filling in placeholders of the templates for verbs and person's names.
From the relationship template, context sentences are generated that explain the social relationships between the speaker and the people mentioned in the input sentence.
In addition, from the sentence template, we create an incorrect or non-honorific sentence and a correct honorific sentence.
The verb conjugates according to the honorific type given when its placeholder is being filled.
We used 23 verbs and 19 names in this experiment.
We chose the verbs which are commonly used in daily conversation.
We also avoid verbs such as \textit{nusumu}(steal) because honorifics cannot usually be applied to disrespectful actions. 
Regarding the names of people, we used the 19 most common family names in Japan in 2022\footnote{\url{https://myoji-yurai.net/prefectureRanking.htm}}.
Finally, a set of the following data is generated from each problem template: context sentences, an incorrect or non-honorific sentence, and a correct sentence.
\begin{table*}[t]
    \small\centering
    \begin{tabular*}{\textwidth}{ll} 
    \toprule
    \multicolumn{2}{l}{Context: The speaker uses honorifics for Kimura. The speaker and Kimura use honorifics for Takahashi-san.}\\
    \multicolumn{2}{l}{\hspace{1.2cm}$\rightarrow$Speaker<Kimura<Takahashi}\\
    \midrule
    Translation & \textit{Takahashi says ``Kimura is going home.''}\\
    Source & Takahashi-san ga ``Kimura ga okaerininaru (=\textit{go-home-\textsc{sh}})'' to ossharu.\\   
    Target & Takahashi-san ga ``Kimura ga kaeru (=\textit{go-home})'' to ossharu. \\
    Model's Prediction & Takahashi-san ga ``Kimura ga \underline{okaerininaru} (=\textit{go-home-\textsc{sh}})'' to ossharu. (Not converted)\\
    \bottomrule
    \end{tabular*}
    \vspace{-0.1cm}
    \caption{An example of the errors regarding direct speech. The speech within brackets is made by Takahashi. The verb \textit{kaeru} should not be in a subject honorific form \textit{okaerininaru} because Takahashi does not use honorifics for Kimura, given their relationships.}
    \label{table:cr_direct_speech}
\end{table*}

\section{Experiments}
\label{sec:cr_experiments}
\subsection{Experimental Setup}
\label{sec:cr_experimental_settings}
We evaluate GPT-3 models on the proposed honorific conversion task under two different experimental settings: fine-tuning and prompt learning.
Despite the general expectation of the superior performance of fine-tuning compared to zero-shot prompt learning, no prior research has aimed to evaluate the performance of LLM on honorific conversion in a prompt-based method.
Thus, we compare the scores of these two methods to validate whether the same goes true for honorific conversion. 
For the two settings accordingly, we use \texttt{davinci}~\cite{brown-gpt3-2020} and \texttt{text-davinci-003}~\cite{ouyang_instructgpt_2022}, which are available in the OpenAI API (see Appendix~\ref{sec:cr_appendix_model_details} for details including hyperparameter settings).
\vspace{-0.08cm}
\paragraph{Fine-tuning}
\label{para:cr_fine_tuning}
We fine-tune two models that differ in the training dataset's size to measure how much data are needed to generalize the problems.
\simpletrain{} is used for training and \simpletest{} and \complextest{} for evaluation.
\threetimes{} is a model trained with 117 problems we prepare by generating three data from each problem template, and in the same way, \seventimes{} is trained with 273 problems.
Although our dataset has relatively little data, we consider it enough for the experiments because the minimum dataset size for fine-tuning GPT-3 is ``a few hundred.''\footnote{\url{https://platform.openai.com/docs/guides/fine-tuning/preparing-your-dataset}}
As shown in Figure~\ref{figure:cr_honorific_conversion}, the input is a concatenation of condition sentences and an incorrect sentence, and the output is a proper honorific sentence.
\vspace{-0.08cm}
\paragraph{Prompt Learning}
\label{para:cr_prompt_learning}
GPT-3 is known for zero-shot learning, solving some tasks given only a natural language description as a prompt. 
In addition to the input text used for fine-tuning, we include a task description in the input prompt (see Appendix~\ref{sec:cr_appendix_prompt_example}).
\vspace{-0.08cm}
\paragraph{Evaluation}
\label{para:cr_evaluation}
We manually calculate the percentage of correct sentences generated by a model.
In this experiment, we regard the output as correct if the verb conjugates to one of the possible honorific forms.
We also ignore mistakes unrelated to verb conjugation (e.g., adding a comma in a natural position). 
We create test datasets using the same problem templates and vocabulary as the training datasets.
\simpletest{} contains 108 examples, and \complextest{} has 408 examples (see Appendix~\ref{sec:cr_appendix_test_dataset}).
\subsection{Results}
\label{sec:cr_results}
Table~\ref{table:cr_evaluation_results} shows the scores under all settings of our experiments on the honorific conversion task.
Overall, the fine-tuning scores surpass those of the prompt-based method.
\begin{table}[t]
    \small\centering
    \begin{tabular}{lcccccc} \toprule
    \multicolumn{2}{c}{\multirow{2}{*}{Setting}} & \multirow{2}{*}{Simple}&\multicolumn{4}{c}{Complex}\\
    \cmidrule(lr){4-7}
    & & & CE & SC & IS & DS  \\
    \midrule
    \multirow{2}{*}{FT} & \threetimes{} &  .889 & .230 & .297 & .081 & \textbf{.368} \\
    & \seventimes{} & \textbf{.990} & \textbf{.326} & \textbf{.452} & \textbf{.231} & .293 \\
    \midrule
    PL & \zeroshot{} & .212 & .115 & .174 & .168 & .100 \\
    \bottomrule
    \end{tabular}
    \caption{Evaluation results of the models on our test dataset through honorific conversion. FT refers to fine-tuning, and PL to prompt learning.}
    \label{table:cr_evaluation_results}
    \vspace{-0.2cm}
\end{table}
\subsubsection{Fine-tuning}
\label{sec:cr_fine_tuning}
The scores plummeted when the models were tested with \complextest{} compared to \simpletest{}.
When we increased the data size, the scores increased in most cases, except when tested for problems with direct speech sentences.
In Table~\ref{table:cr_direct_speech}, the model failed to convert a direct speech sentence (\textit{Takahashi-san ga ``Kimura ga okaerininaru'' to ossharu}).
The verb \textit{kaeru} should not be in a subject honorific form (\textit{okaerininaru}) because Takahashi does not use honorifics for Kimura, given their relationships.
However, if the brackets (quotation marks in Japanese, see Appendix~\ref{sec:cr_appendix_templates}) are removed, the sentence (\textit{Takahashi-san ga Kimura ga okaerininaru to ossharu}) becomes an indirect speech sentence and thus becomes proper honorifics.
Based on this characteristic, we suppose that the model applied the same honorific knowledge as indirect speech to direct speech, ignoring the role of brackets.
\subsubsection{Prompt Learning}
\label{sec:cr_prompt_learning}
The scores were relatively higher when tested with \simpletest{} than with \complextest{}, but the scores under all of our settings were lower than 25\%.
We found that the models transferred non-honorific sentences to polite forms in almost all cases by simply changing the last letters of the verbs that end \textit{-suru} into \textit{-shimasu} instead of applying SH or OH.
This conversion is possibly caused by our prompt instructing the models to ``convert to the proper honorific sentence,'' which may include polite forms too.
To validate whether the models use contextual information, we need to construct a prompt that can differentiate SH and OH from polite speech because polite forms are less restricted to social relationships.

\section{Conclusions and Future Work}
\label{sec:cr_conclusions_and_future_work}
In this paper, we introduced an honorific conversion task that requires not only syntactic knowledge but also pragmatic knowledge, such as
social relationships among people.
We constructed a Japanese honorific dataset using problem templates created manually and evaluated the syntactic generalization capacity of GPT-3 models on the task using our dataset.
The experiments showed that the fine-tuned models could solve problems with simple structures but failed to generalize to problems with more complex structures, particularly with direct speech.
Regardless of the sentence structure, the prompt-based models did not successfully solve the problems with our current prompt setting.

In future work, we plan to expand our dataset to include more diverse Japanese honorific expressions, such as predicates other than verbs or honorific prefixes attached to nouns.
For the prompt-based experiments, we evaluated the models using zero-shot learning.
It would be valuable to test them using few-shot learning by including simple examples in the prompts.

We conducted experiments by explicitly providing information about social relationships.
We will also continue to seek data construction methods to extract such information from the corpora, although we did not apply these corpora-based methods in this paper.
\clearpage
\section*{Limitations}
\label{sec:cr_limitations}
We discuss two limitations of this research in this section.
First, this research focuses on Japanese honorifics with grammatical rules of verb conjugation, which we can judge whether the honorific conversion is correct based on social relationships and sentence structures created in a controlled manner.
Japanese honorifics have more expressions based on linguistic phenomena that we did not include in our templates, such as noun honorifics (e.g., \textit{ofutagata}, a polite and formal way of saying ``the two people'').
Creating templates for noun honorifics requires more detailed settings because they are based on information on context other than social relationships.
Second, GPT-3 is the only language model evaluated on our honorific conversion task. 
This research aims to analyze how capable the well-known, high-performing GPT-3 is of generalizing Japanese honorific sentences and not to explore which existing LLM can achieve the best performance in honorific conversion.

\section*{Acknowledgements}
We thank three anonymous reviewers for their helpful comments and suggestions. 
We would also like to thank Anirudh Reddy Kondapally, Tomoki Sugimoto, Tomoya Kurosawa, and the other laboratory members for their helpful advice.
This work was supported by JSPS KAKENHI Grant Number JP20K19868.

\bibliography{anthology,custom}
\bibliographystyle{acl_natbib}

\appendix
\begin{table*}[t]
    \caption{Examples from the existing honorific datasets.}
    \small\centering
    \begin{tabular}{lll} 
    \toprule
    Original & Converted & Label \\
    \midrule\midrule
    朝ごはんはトーストにバターとべジ & 朝ごはんはトーストにバターとべジ & 変換:謙譲語 \\
    マイトを薄くぬって食べました。&マイトを薄くぬっていただきました。& \\
    (I had toast for breakfast with a &(I \textbf{had\_\textsc{oh}} toast for breakfast with a&(Converted: OH) \\
    thin layer of butter and Vegemite.)& thin layer of butter and Vegemite.) \\
    \midrule
    そして10時くらいに、喫茶店でレー & そして10時くらいに、喫茶店でレー & 無変換 \\
    シャルとジョノサンとベルに会いました。& シャルとジョノサンとベルに会いました。 & \\
    (Then, at around 10:00, I met Rachel, & (Then, at around 10:00, I met Rachel, & (Not converted) \\
    Jonathan, and Belle at a coffee shop.)& Jonathan, and Belle at a coffee shop.) & \\
    \bottomrule
    \end{tabular}
    \vspace{0.2cm}
    \subcaption*{\cite{matsumoto_honorifics-style-transfer_2022}}
    \medskip
    \includegraphics[width=0.95\textwidth]{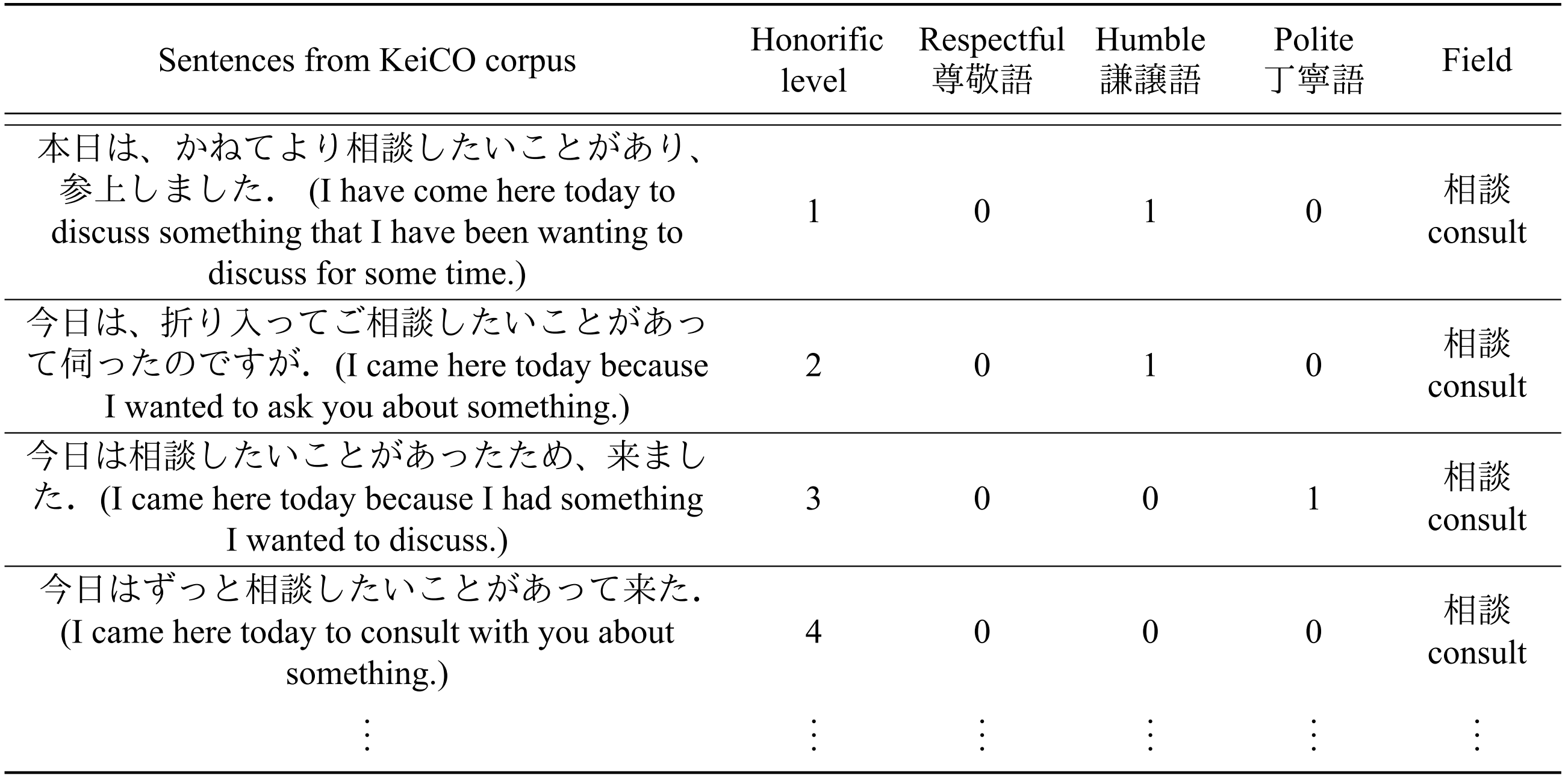}
    \subcaption*{\cite{liu_construction_2022}}
    \label{table:cr_existing_datasets}
\end{table*}
\section{Existing Datasets}
\label{sec:cr_appendix_existing_datasets}
Table~\ref{table:cr_existing_datasets} shows examples of existing Japanese honorifics datasets.
\section{Templates}
\label{sec:cr_appendix_templates}
Table~\ref{table:cr_templates} shows examples of templates we created.
Table~\ref{table:cr_indirect_direct} shows examples of indirect and direct speech in Japanese.
\section{Model Details}
\label{sec:cr_appendix_model_details}
\texttt{davinci} is the largest model among the ones provided for fine-tuning, and \texttt{text-davinci-003} is trained by reinforcement learning on human feedback and aimed at being used with prompt learning.
\paragraph{Hyperparameters}
For fine-tuning GPT-3, \texttt{n\_epochs} is set to 2.
For text generation, we set \texttt{max\_tokens} to 50 and \texttt{temperature} to 0.
\paragraph{Pre-training Data of GPT-3}
GPT-3 can input and output Japanese texts because some of its pre-training datasets (Common Crawl, WebText) contain Japanese texts, although the proportion of Japanese texts is not clarified.
\section{Test Dataset}
\label{sec:cr_appendix_test_dataset}
Within the \complextest{} dataset, 156 data have center embedding, 252 for scrambling, 160 for indirect speech, and 160 for direct speech. 
Scrambling and center embedding can not appear in one problem data; the same goes for indirect and direct speech.
\section{Prompt Example}
\label{sec:cr_appendix_prompt_example}
\begin{figure}[t]
    \small\centering
    \begin{screen}
    \textbf{以下の文はあなたの発言です。人物間の敬語の条件を踏まえて、敬語が不十分かそれらを誤って使っている場合は正しい敬語に変換してください。(The following sentence is your speech. Given the condition of usage of honorifics between people, convert the sentence to the proper honorific one if it contains wrong or insufficient honorifics.)}\\
    ===\\
    敬語の条件:あなたは田中に敬語を使います。(Condition: You use honorifics for Tanaka.)\\
    田中が受け取る (Tanaka receives) ->
    \end{screen}
    Correct output: 田中がお受け取りになる (Tanaka receives-\textsc{sh})
    \caption{An example of the prompt used for zero-shot learning. The bold text is a task description.}
    \label{figure:cr_prompt_example}
\end{figure}
Figure~\ref{figure:cr_prompt_example} shows an example of our prompt used for prompt learning.
\clearpage
\begin{table*}[t]
    \small\centering
    \begin{tabular}{llll} 
    \toprule
    Relationship template & Honorific type & \wraptext{Sentence template\\and an example of created correct sentence} & Structure type\\ 
    \midrule\midrule
    speaker=actor\_1=target\_1 &
    \wraptext{v\_ni\_1 $\rightarrow$ NA}&
    \wraptext{actor\_1 \textit{ga} target\_1 \textit{ni} v\_ni\_1。\\ 
            actor\_1 \nominative{} target\_1 \dative{} v\_ni\_1。\\
            Sasaki ga Saito ni au。\\
            (Sasaki meets Saito.)} &
            \simple{}
    \\ \midrule
    
    speaker=target\_1<actor\_1&
    \wraptext{v\_ni\_1 $\rightarrow$ SH}& 
    \wraptext{actor\_1 \textit{ga} target\_1 \textit{ni} v\_ni\_1。\\ 
                actor\_1 \nominative{} target\_1 \dative{} v\_ni\_1。\\
                Takahashi-kyoju ga Kimura ni o-ai-ninaru。\\
                (Prof. Takahashi meets Kimura.)}&
                \simple{}
    \\ \midrule
    
    speaker=target\_1<actor\_1 &
    \wraptext{v\_o\_1 $\rightarrow$ SH}& 
    \wraptext{actor\_1 \textit{ga} target\_1 \textit{o} v\_o\_1。\\
            actor\_1 \nominative{} target\_1 \accusative{} v\_o\_1。\\
            Kimura-hakase ga Yamada o shokai-nasaru。\\
            (Dr. Kimura introduces Yamada.)}&
            \simple{}
    \\ \midrule
    
    speaker=actor\_1=target\_1 & 
    \wraptext{v\_ni\_1 $\rightarrow$ NA}& 
    \wraptext{target\_1 \textit{ni} actor\_1 \textit{ga} v\_ni\_1。 \\
            target\_1 \dative{} actor\_1 \nominative{} v\_ni\_1。\\
            Kimura ni Yamamoto ga kanshasuru。\\
            (Yamamoto thanks Kimura.)}&
            \complex{} (SC)
    \\ \midrule
    
    speaker=actor\_1=actor\_2 &
    \wraptext{v\_to\_1 $\rightarrow$ NA \\ v\_single\_2 $\rightarrow$ NA}&
    \wraptext{actor\_1 \textit{ga} ``actor\_2 \textit{ga} v\_single\_2'' \\ \qquad         \textit{to} v\_to\_1。\\
            actor\_1 \nominative{} ``actor\_2 \nominative{} v\_single\_2''  \\ \qquad \citative{} v\_to\_1。\\
            Itoh ga ``Matsumoto ga iku''  to iu。\\
            (Itoh says ``Matsumoto goes.'' )}&
            \complex{} (DS, CE)
    \\ \midrule
    
    speaker<actor\_2<actor\_1 &
    \wraptext{v\_to\_1 $\rightarrow$ SH \\ v\_single\_2 $\rightarrow$ NA}&
    \wraptext{``actor\_2 \textit{ga} v\_single\_2''  \textit{to} \\ \qquad actor\_1 \textit{ga} v\_to\_1。\\
            ``actor\_2 \nominative{} v\_single\_2''  \citative{} \\ \qquad actor\_1 \nominative{} v\_to\_1。\\
            ``Kimura-sensei ga uketoru'' to \\ \qquad Kato-hakase ga o-kangae-ninaru。\\
            (Dr. Kato considers, \\ \qquad ``Kimura-sensei will receive it.'' )}&
            \complex{} (DS, SC)
    \\ \midrule
    
    speaker<actor\_2<actor\_1&
    \wraptext{v\_to\_1 $\rightarrow$ SH \\ v\_single\_2 $\rightarrow$ SH}&
    \wraptext{actor\_2 \textit{ga} v\_single\_2 \textit{to} \\ \qquad actor\_1 \textit{ga} v\_to\_1。\\
            actor\_2 \nominative{} v\_single\_2 \citative{} \\ \qquad actor\_1 \nominative{} v\_to\_1。\\
            Kimura-sensei ga o-uketori-ninaru to \\ \qquad Kato-hakase ga o-kangae-ninaru。\\
            (Dr. Kato considers that \\ \qquad Kimura-sensei will receive it.)}&
            \complex{} (IS, SC)
    \\ \bottomrule
    \end{tabular}
    \caption{Examples of problem templates. NA in the honorific type section means no honorific needs to be applied to a verb. SC=scrambling, CE=center embedding, DS=direct speech, IS=indirect speech}
    \label{table:cr_templates}
\end{table*}
\begin{table*}[t]    
    \centering
    Social relationships: Speaker<Taro=Hanako \\
    \vspace{1.0mm}
    \begin{tabular}{ll} 
    \toprule    
    Indirect speech & 
    \wraptext{Taro-\textbf{san}-ga \textbf{irasshatta} to Hanako-san-ga itta. \\
    Taro-\textbf{\textsc{hon}}-\nominative{} \textbf{came-\textsc{sh}} \citative{} Hanako-\textsc{hon}-\nominative{} said.} 
    \\ \midrule
    Direct speech & 
    \wraptext{「Taro-ga \textbf{kita}」 to Hanako-san-ga itta. \\ Taro-\nominative{} \textbf{came} \citative{} Hanako-\textsc{hon}-\nominative{} said.}
    \\ \bottomrule
    \end{tabular}
    \caption{Examples of indirect speech and direct speech in Japanese. Indirect speech is the citation of someone's speech without quotation marks (brackets), and direct speech is the one with them.
    In the example of indirect speech, subject honorifics are applied to Taro's name (\textit{-san}) and his action (\textit{irasshatta}) to express the speaker's respect for him. In contrast, the sentence within brackets is written without any honorifics in direct speech. Hanako does not use honorifics for Taro's actions according to their social relationships, so the quoted sentence is what Hanako said, and no honorifics from the speaker's view of the entire sentence are reflected.}
    \label{table:cr_indirect_direct}
\end{table*}

\end{document}